\documentclass[sigconf]{acmart}
\usepackage{epsfig}
\usepackage[ruled]{algorithm2e}
\usepackage{threeparttable}
\usepackage{adjustbox} 
\usepackage{booktabs}
\usepackage{listings}
\AtBeginDocument{%
  \providecommand\BibTeX{{%
    \normalfont B\kern-0.5em{\scshape i\kern-0.25em b}\kern-0.8em\TeX}}}

\usepackage{amsmath,amsfonts,amscd,amsthm,xspace}
\usepackage{stfloats}
\usepackage{balance}
\usepackage{tabularx}
\usepackage{multirow, multicol}
\usepackage{url}

\usepackage{caption}
\usepackage{subcaption}

\SetKwProg{Fn}{function}{}{}

\copyrightyear{2021}
\acmYear{2021}
\setcopyright{acmlicensed}
\acmConference[SAC '21]{The 36th ACM/SIGAPP Symposium on Applied Computing}{March 22--26, 2021}{Virtual Event, Republic of Korea}
\acmBooktitle{The 36th ACM/SIGAPP Symposium on Applied Computing (SAC '21), March 22--26, 2021, Virtual Event, Republic of Korea}
\acmPrice{15.00}
\acmDOI{10.1145/3412841.3441957}
\acmISBN{978-1-4503-8104-8/21/03}

\begin{document}

\title[Explaining a Neural Att. Model for ABSC Using Diagnostic
Classification]{Explaining a Neural Attention Model for Aspect-Based Sentiment Classification Using Diagnostic Classification}




\author{Lisa Meijer}
\affiliation{  \institution{Delft University of Technology}
  \city{Delft}
  \country{the Netherlands}
  \streetaddress{Mekelweg 5}
  \postcode{2628 CD}
}
\email{lhmeijer@tudelft.nl}

\author{Flavius Frasincar}
\orcid{0000-0002-8031-758X}
\affiliation{ 
  \institution{Erasmus University Rotterdam}
  \streetaddress{Burgemeester Oudlaan 50}
  \city{Rotterdam}
  \country{the Netherlands}
  \postcode{3062 PA}
}
\email{frasincar@ese.eur.nl}

\author{Maria Mihaela Tru\c{s}c\u{a}}
\affiliation{  \institution{Bucharest Univ. of Economic Studies}
  \city{Bucharest}
  \country{Romania}
  \postcode{010374}
  \streetaddress{Piata Romana 6}
}
\email{maria.trusca@csie.ase.ro}

%

\begin{abstract}
Many high performance machine learning models for Aspect-Based Sentiment Classification (ABSC) produce black box models, and therefore barely explain how they classify a certain sentiment value towards an aspect. In this paper, we propose explanation models, that inspect the internal dynamics of a state-of-the-art neural attention model, the LCR-Rot-hop, by using a technique called Diagnostic Classification. Our diagnostic classifier is a simple neural network, which evaluates whether the internal layers of the LCR-Rot-hop model encode useful word information for classification, i.e., the part of speech, the sentiment value, the presence of aspect relation, and the aspect-related sentiment value of words. We conclude that the lower layers in the LCR-Rot-hop model encode the part of speech and the sentiment value, whereas the higher layers represent the presence of a relation with the aspect and the aspect-related sentiment value of words.
\end{abstract}

%



\maketitle


\section{Introduction}
Aspect-based Sentiment Classification (ABSC) is a subtopic of the Textual Classification task whose main aim is the assignment of sentiment orientations (positive, negative, or neutral) at the aspect-level of the entity of interest. In the recent years, many models have been designed for ABSC, among which machine learning together with its deep learning subset have proved to be the most effective. However, the performance of machine learning models for the ABSC is usually traded off for a high level of uncertainty and vagueness in the logic of models \cite{liu2015sentiment}. The reason behind the unsuccess in clarifying the model's inner working \cite{Strobelt2018} is especially by the large numbers of parameters that need to be optimised. In addition to the benefits to businesses, the regulation of General Data Protection Regulation (GDPR) \cite{voigt2017eu} has shown us that it has become a societal need to make machine learning models more understandable, and to transform black boxes into transparent, white boxes. 

In this paper, we use a state-of-the-art ABSC model, the Left-Center-Right Rotatory Attention model with multiple hops (LCR-Rot-hop), proposed by \citet{Wallaart2018}. Given the LCR-Rot-hop model, it is unclear which input information gets encoded while processing an opinion, and whether the model has any understanding about which word(s) determine(s) the sentiment polarity towards an aspect. Therefore, our research aim is to shed light on the inner working of the LCR-Rot-hop model by means of Diagnostic Classification \cite{DBLP:journals/jair/HupkesVZ18}. Precisely, we use Diagnostic Classification to examine whether the internal layers of the LCR-Rot-hop model encode for each word of an opinion the \textit{part of speech}, the \textit{presence of aspect relation}, the \textit{sentiment value}, and the \textit{aspect-related sentiment value} information that it is useful for the current classification task. By means of the LCR-Rot-hop model, we provide a methodology to understand the inner information process specific to the ABSC problem that can be easily applied to any other neural network designed for this purpose. Our implementation can be found at \url{https://github.com/lhmeijer/ABSCEM}. 

The structure of this paper is as follows. In Section \ref{sec:related_work}, we discuss the relevant literature of ABSC using neural networks, and of Diagnostic Classification. Section \ref{sec:data} gives an overview of the used dataset. Section \ref{sec:method} gives a description of the LCR-Rot-hop model, and our proposed diagnostic classifiers, which we evaluate in Section \ref{sec:evaluation}. In Section \ref{sec:conclusion} we present our conclusion and future work. 

\section{Related Work}
\label{sec:related_work}
The following Section \ref{sec:absc} and Section \ref{sec:diagnostic} review the literature dedicated to ABSC and Diagnostic Classification, respectively.

\subsection{Aspect-Based Sentiment Classification}
\label{sec:absc}
Since a writer could express multiple opinions towards different specific aspects in one document or sentence, only looking at a document or a sentence level as in Sentiment Classification is often insufficient \cite{liu2015sentiment}. Therefore, we focus on ABSC, a finer-grained sentiment classification, which computes the sentiment for an aspect of an entity of interest, at the sentence level \cite{schouten2015survey}.



Nowadays Machine Learning approaches are popular ABSC models. Frequently, these models are based on neural networks, such as Long Short-Term Memory (LSTM) networks \cite{Hochreiter1997} and neural attention models \cite{Liu2018}. \citet{Tang2016} proposed a target-dependent LSTM (TD-LSTM) approach, which models the relatedness of an aspect with its context words, and selects the relevant parts from the context to deduce the sentiment value towards the aspect. \citet{Liu2018} developed the Content Attention Based Aspect-Based Sentiment Classification (CABASC) model, which consists of two attention modeling mechanisms, one at the sentence level and one at the context level. While the sentence level content attention mechanism considers the whole meaning of the full sentence, the context attention mechanism takes the correlations between the words and the aspect into account. 

\citet{Zheng2018} suggested a Left-Center-Right separated neural network with a rotatory attention mechanism (LCR-Rot) to effectively represent an aspect composed of multiple words, and to utilize the interaction between the aspect and its context. Additionally, \citet{Wallaart2018} proposed the LCR-Rot-hop model as an extension of the the work of \citet{Zheng2018}. The LCR-Rot-hop model iterates the rotatory attention mechanism to better indicate the relation between the aspect and the context of an opinion. Given that the LCR-Rot-hop model is a state-of-the-art approach to ABSC with higher rates of effectiveness than TD-LSTM, LCR-Rot, and CABASC models, we only consider this model for our diagnostic analysis.



\subsection{Diagnostic Classification}
\label{sec:diagnostic}
Although, neural networks are good at predicting the sentiment value of an opinion, they are not highly transparent predictors. Therefore, \citet{DBLP:journals/jair/HupkesVZ18} proposed a Diagnostic Classifier that attempts to predict information from the (hidden) states of a neural network. If this information is predicted accurately, it indicates that the information is indeed computed or represented by the network at the level of the given hidden states.

The idea of a diagnostic classifier is based on a prediction task proposed by \citet{adi2016}. They trained a classifier to predict specific sentence properties (e.g., number of words, word context, and word order) based on the vector representations obtained from LSTM autoencoders or simple continuous bag-of-words representations. \citeauthor{adi2016} stated that a property of a sentence is not encoded in the representation, if we cannot train a classifier to predict this property.

Initially, \citet{DBLP:journals/jair/HupkesVZ18} proposed a diagnostic classifier to study how recurrent neural networks process hierarchical structures by using an arithmetic language, but \citeauthor{DBLP:journals/jair/HupkesVZ18} mainly showed that diagnostic classification is a useful technique for opening up the black box of neural networks. For instance, \citet{DBLP:conf/emnlp/JumeletH18} used the diagnostic classification to predict whether a word in a sentence is inside the licensing scope. \citeauthor{DBLP:conf/emnlp/JumeletH18} concluded that their neural language model is able to encode this information.

\citet{DBLP:conf/acl/BelinkovDDSG17} evaluated the representation quality of neural machine translation models on part-of-speech and morphological tagging in various languages by using a neural classifier. \citet{DBLP:conf/acl/BelinkovDDSG17} concluded that lower layers of the neural machine translation models are better at capturing morphology than the higher layers.

\section{Specification of the Data}
\label{sec:data}
In this paper, we use the widely employed SemEval-2016 Task 5 Subtask 1 dataset \cite{Pontiki2016} to train, and evaluate the LCR-Rot-hop model. The SemEval-2016 dataset consists of hundreds of restaurant reviews. Each review is divided into various sentences, and each sentence has one or multiple opinions. 

Figure \ref{fig:exampleSen} presents a review in the XML markup language. This sentence includes three opinions. For each opinion, a \texttt{polarity} (sentiment value) is specified, which expresses the sentiment of the reviewer towards the predefined aspect or \texttt{target}. Moreover, an aspect \texttt{category} is stated, which classifies the \texttt{target}.
\definecolor{maroon}{rgb}{0.5,0,0}
\definecolor{darkgreen}{rgb}{0,0.5,0}
\lstdefinelanguage{XML}
{
  basicstyle=\ttfamily,
  morestring=[s]{"}{"},
  morecomment=[s]{?}{?},
  morecomment=[s]{!--}{--},
  commentstyle=\color{darkgreen},
  moredelim=[s][\color{black}]{>}{<},
  moredelim=[s][\color{red}]{\ }{=},
  stringstyle=\color{blue},
  identifierstyle=\color{maroon},
  frame = single
}
\begin{figure}[h]
\vspace{-3mm}
\small
\begin{lstlisting}[language=xml]
<Review rid="404464">
<sentences>
<sentence id="404464:0">
<text>Thalia is a beautiful restaurant with 
beautiful people serving you, but the food 
doesn't quite match up.</text>
<Opinions>
<Opinion target="people" 
category="SERVICE#GENERAL" polarity="positive" 
from="48" to="54"/>
<Opinion target="food" category="FOOD#QUALITY" 
polarity="negative" from="76" to="80"/>
<Opinion target="Thalia" 
category="AMBIENCE#GENERAL" polarity="positive" 
from="0" to="6"/>
</Opinions>
</sentence>
</sentences>
</Review>    
\end{lstlisting}
\caption{A sentence from the SemEval-2016 dataset}
\label{fig:exampleSen}
\vspace{-4mm}
\end{figure}

Since the model relies on the presence of explicitly stated opinions, we remove all implicitly stated opinions from the dataset. Moreover, we replace all long dashes ($-$) by a short dash (-), and all textual apostrophes (’) by a digital typesetting (\textquotesingle) to encode these similar characters in the same way. We use the Stanford CoreNLP package \cite{Manning14thestanford} to tokenize and lemmatize all the words within a sentence. 

To symbolise the meaning of a word, such that we can use it as input in our model, we represent a lemmatized word as a point in a 300 dimensional semantic space. We use the pre-trained algorithm GloVe \cite{pennington2014glove} to obtain these vector representations (embeddings). If words do not appear in the GloVe vocabulary, we randomly initialise these words by using a normal distribution $N(0,0.052)$ \cite{Wallaart2018}.

\section{Method}
\label{sec:method}
Below, we first give an overview of the LCR-Rot-hop model in Section \ref{method_lcr_rot}, and then present the diagnostic classification in Section \ref{method_diagnostic}.

\subsection{Neural Rotatory Attention model with multiple hops} \label{method_lcr_rot}
According to the LCR-Rot-hop model \cite{Wallaart2018}, we initially split the sentences in a left, a center (target), and a right part depending on the position of the target. The left part consists of $L$ words, the right part has $R$ words, and the target phrase consists of $T$ words belonging to the aspect. 


The model starts with three bi-directional Long-Short-Term-Memory (Bi-LSTM) networks for each of the three parts of the input sentence (left context, target content, and right context). These Bi-LSTMs use, as input, the embeddings of the words in the left context $[e_1^l, ..., e^l_L]$, the embeddings of the words in the target phrase $[e_1^t, ..., e^t_T]$, and the embeddings of the words in the right context $[e_1^r, ..., e^r_R]$ to compute the corresponding hidden states $[h_1^l, ..., h^l_L]$ for the left context, $[h_1^t, ..., h^t_T]$ for the target phrase, and $[h_1^r, ..., h^r_R]$ for the right context.

To capture the most sentiment indicative words in a sentence, we apply a rotatory attention mechanism to all hidden states. First, we use an attention scoring function for both the left $f(h_i^l,r^{t_p})$ and right $f(h_i^r,r^{t_p})$ contexts to represent the most sentiment indicative words in the left and right contexts, respectively. Both the attention scoring functions are defined with respect to their hidden states, and the average target representation $r^{t_p}$, calculated by feeding the target hidden states into an average pooling layer. Given the symmetry of the rotatory attention mechanism, we present the computational process only for one context. As a result the attention scoring function for the left context is defined as follows:
\begin{equation}
f(h_i^l, r^{t_p}) = \tanh(h_i^l \times W_c^l \times r^{t_p} + b_c^l)
\label{eq:1}
\end{equation} 
where $W_c^l$ is a weight matrix, and $b_c^l$ is a bias term. We feed these context attention scores $f(h_i^l,r^{t_p})$ and $f(h_i^r,r^{t_p})$ into a softmax function for normalization. With the normalised attention scores $\alpha_i^l$ and $\alpha_i^r$, we compute the left $r^l$ and right $r^r$ context representations. Considering the left context again, the $r^l$ vector is computed as:
\begin{equation}
r^l = \sum_{i=1}^{L} \alpha^l_i \times h^l_i
\label{eq:2}
\end{equation}
Second, we use these context representations $r^l$ and $r^r$, and an attention scoring function for both the left-aware target phrase $f(h_i^t,r^l)$ and the right-aware target phrase $f(h_i^t,r^r)$ to represent the most sentiment indicative words in the target phrase. The attention scoring function defined with respect to the left context is:
\begin{equation}
f(h_i^t,r^l) = \tanh(h_i^t \times W_t^l \times r^l + b_t^l)
\label{eq:3}
\end{equation}
where $W_t^l$ is a weight matrix and $b_t^l$ is a bias term. These target attention scores $f(h_i^t, r)$ are as well fed into a softmax function to obtain the normalised attention scores $\alpha_i^{t_l}$ and $\alpha_i^{t_r}$. We use these scores to compute the left-aware target $r^{t_l}$ and right-aware target $r^{t_r}$ representations. Similar to the $r^l$ vector defined above (eq. \ref{eq:2}), the $r^{t_l}$ representation is computed as:
\begin{equation}
r^{t_l} =  \sum_{i=1}^{T}\alpha^{t_l}_i \times h^t_i
\label{eq:4}
\end{equation}
Next, we repeat the above mentioned steps of applying the rotatory attention mechanism by feeding the left- and right-aware target representations into the left and right attention functions, respectively, instead of the average target representation ($r^{t_p}$), and we compute again all context and target representations. Moreover, the rotatory attention mechanism is repeated over multiple iterations to better represent the relation between the aspect and its context. 

At the end of the iterative rotatory attention mechanism, we concatenate the left-context representation $r^l$, the two side-target representations, $r^{t_l}$ and $r^{t_r}$, and the right-context representation $r^r$, to obtain the sentence representation $s$. Further, the new representation of the input sentence is converted by a dense layer with a softmax activation function to compute the sentiment prediction vector with a score for each of the three \textit{sentiment values}: positive, negative, and neutral.

The LCR-Rot model is trained in a supervised manner by minimising the cross-entropy loss function with an $L_2$-norm regularisation term. To update the weights and biases we use stochastic gradient descent with momentum, and to prevent overfitting we apply the dropout technique to all hidden layers. For a more in-depth discussion of the training process, we refer to \cite{Wallaart2018}.

\subsection{Diagnostic Classification} \label{method_diagnostic}
\label{sec:diagclass}
Diagnostic Classification is based on the idea that if a model is keeping track of certain information, we should be able to extract this information from the internal layers of the model. We propose several diagnostic classifiers to examine whether the internal layers of the LCR-Rot-hop model are keeping track of the \textit{part of speech information}, the presence of a \textit{relation with the aspect}, the \textit{sentiment value}, and the \textit{sentiment value related to the aspect}, all information deemed useful at the word level for ABSC. If the diagnostic classifiers are able to predict accurately, it indicates that the considered information is indeed encoded by the layers of the LCR-Rot-hop model.

Since our diagnostic classifiers try to predict from the internal layers of the LCR-Rot-hop model, we first train the LCR-Rot-hop model on the SemEval-2016 training dataset to obtain the internal layers of all words in the opinions of the training and test dataset. For each word $i$, we obtain its embedding ($e_i$); its hidden state ($h_i$); and its $j\in\{1,... ,n\}$ context representations ($r_{ij}$), where $j$ represents the hop number. Since the context representation $r$ is computed as a single value for a group of words (eq. \ref{eq:2}), the $r_{ij}$ vector representation at the word-level is defined as follows:
\begin{equation}
r_{ij} =  \alpha_{ij} \times h_i
\label{eq:5}
\end{equation}
where $\alpha_{ij}$ is either the left attention score ($\alpha^{l}_{ij}$) or the right attention score ($\alpha^{r}_{ij}$) for the $j$ iteration of the rotatory attention mechanism. In order to predict from the hidden layers ($e_i, h_i, r_{i1}, ..., r_{in}$) the \textit{part-of-speech}, the \textit{presence of a relation with the aspect}, the \textit{sentiment value}, and the \textit{sentiment value in relation to the aspect}, we define below the information that makes up the class labels.

To predict the \textit{part-of-speech} tags, we categorise the words as: noun, verb, adjective (adj.), or adverb (adv.). If a word is none of those four, we categorize it as remaining \textit{part-of-speech} (rem.). We simply extract these labels from each individual word by using the Stanford CoreNLP package \citep{Manning14thestanford}. Table \ref{tab:pos} represents the number of words per \textit{part-of-speech} class. The diagnostic classifiers are trained only on the correctly predicted instances of the training data because it does not make sense to use the incorrectly predicted training data that might cause the learning of inconsequential patterns. Therefore, we choose to not present the distribution per classes of the words in the training instances that are incorrectly predicted by the LCR-Rot-hop in Tables \ref{tab:pos}-\ref{tab:apect_sentiment}.

\begin{table}[h!]
\caption{Number of words per \textit{part-of-speech} class.}
\centering
\footnotesize
\begin{tabular}{l p{0.06\linewidth} p{0.06\linewidth} p{0.06\linewidth} p{0.06\linewidth} p{0.08\linewidth}}
\toprule
& Noun & Verb & Adj. & Adv. & Rem.\\
\midrule
Correctly predicted training set & 4586 & 3631 & 2058 & 4661 & 14148 \\
Correctly predicted test set & 1211 & 1099 & 569 & 2338 & 4848 \\
Incorrectly predicted test set & 278 & 182 & 129 & 264 & 908 \\
\bottomrule
\end{tabular}
\label{tab:pos}
\end{table}

To predict the \textit{presence of a relation} with the aspect, we label a word as ``Yes'' if it is related to the aspect, and as ``No'', if it is not related to the aspect. We determine the existence of a relation by using the Stanford CoreNLP Dependency Parser \citep{Manning14thestanford}, and the ontology described by \citet{Schouten2017} that connects aspects to words. A word is related to the aspect in the opinion if there exists a dependency between them, or if they are connected according to the ontology. The number of words per \textit{presence of a relation} class is given in Table \ref{tab:rel}. 
\begin{table}[h!]
\caption{Number of words per \textit{presence of a relation} class.}
\centering
\footnotesize
\begin{tabular}{l c c}
\toprule
& Yes & No\\
\midrule
Correctly predicted training set & 6220 & 22864\\
Correctly predicted test set & 2070 & 7995\\
Incorrectly predicted test set & 364 & 1397\\
\bottomrule
\end{tabular}
\label{tab:rel}
\end{table}

To predict the \textit{sentiment value}, we assign each word to the class label ``Positive'', if the word has a positive \textit{sentiment value}, ``Negative'', if the word has a negative \textit{sentiment value}, and ``No Sentiment'', if we cannot define the \textit{sentiment value}. Table \ref{tab:word_sentiment} represents the classes' frequencies in the data sets. We specify the \textit{sentiment value} of a word by using the ontology described by \citet{Schouten2017}, and NLTK SentiWordNet \citep{Bird2009}. The ontology classifies a word positive (negative), if the word has a superclass \emph{Positive} (\emph{Negative}). If the ontology is not able to classify the \textit{sentiment value}, we use SentiWordNet \citep{Bird2009} that defines the sentiment score of the word's most frequently used form. A word has a positive \textit{sentiment value}, if the positive sentiment score is larger than the negative sentiment score, and vice versa. We classify the word as ``No Sentiment'', if both the ontology and SentiWordNet are not able to identify the \textit{sentiment value}. Table \ref{tab:word_sentiment} lists the number of words per \textit{sentiment value} class.
\begin{table}[h!]
\caption{Number of words per \textit{sentiment value} class.}
\centering
\footnotesize
\begin{tabular}{l c c c}
\toprule
& Positive & Negative & No Sentiment\\
\midrule
Correctly predicted training set & 5583 & 1774 & 21727\\
Correctly predicted test set & 1677 & 489 & 7899\\
Incorrectly predicted test set & 264 & 120 & 1377\\
\bottomrule
\end{tabular}
\label{tab:word_sentiment}
\end{table}

To predict the \textit{sentiment value of a word related to the aspect} in the opinion, we classify a word as ``Positive'' if it has both a positive \textit{sentiment value} and a relation towards the aspect, as ``Negative'' if it has both a negative \textit{sentiment value} and a relation towards the aspect, and as ``No Aspect Sentiment'' if the \textit{sentiment value} or the relation are undefined. We specify the \textit{sentiment value} in the same way, as we did for predicting the \textit{sentiment value}, and we determine the existence of a relation in the same way, as we did for predicting the \textit{presence of a relation}. Table \ref{tab:apect_sentiment} gives the number of words per \textit{aspect-related sentiment value} class.

\begin{table}[h!]
\caption{Number of words per \textit{aspect-related sentiment value} class.}
\centering
\footnotesize
\begin{tabular}{l c c c}
\toprule
& Pos. & Neg. & No Aspect Sentiment\\
\midrule
Correctly predicted training set & 1425 & 439 & 27220\\
Correctly predicted test set & 450 & 91 & 9524\\
Incorrectly predicted test set & 57 & 38 & 1666\\
\bottomrule
\end{tabular}
\label{tab:apect_sentiment}
\end{table}

Since the number of words per class label for any information is rather unequally distributed, we run the LCR-Rot-hop and select sub-samples from the correctly predicted opinions of the SemEval-2016 training set to train the diagnostic classifiers. We randomly draw $\min(q_c, q^{mean})$ words per class, where $q_c$ is the number of word with class label $c$, and $q^{mean}$ is the mean of all $q_c$'s except from the class that has the greatest number of words. The reason behind this configuration is the need to reduce the differences between the classes in terms of size.

\begin{figure}[t!]
	\centering
	\includegraphics[width=0.46\textwidth]{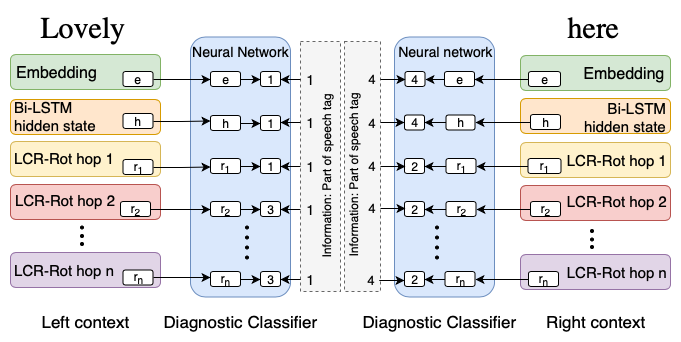}
	\caption{Diagnostic classifier applied for the detection of \textit{part-of-speech} of the words ``Lovely" and ``here" }
	\label{fig:diagnostic_classifier}
\end{figure}

After composing multiple training sets from the internal layers of the randomly selected words, and their corresponding class labels, we train the various diagnostic classifiers on their related training set. Figure \ref{fig:diagnostic_classifier} gives an illustration of all $2+n$ diagnostic classifiers, which try to predict the \textit{part-of-speech} of the words ``Lovely'' and ``here'' from their various hidden layers ($e, h, r_{1}, ..., r_{n}$). The words ``Lovely'' and ``here'' represent the left and right contexts of the imaginary opinion ``Lovely place here", where the ``place" word is the aspect. ``Lovely” is an adjective, and should be predicted as 1, while ``here” has a remaining tag, and should be predicted as 4 (where the class labels are ordered from 0 to 4). Since we select a sub-sample to train the diagnostic classifiers on, a single run is insufficient to draw a conclusion. Therefore we train our diagnostic classifiers on 10 subsets of the correctly predicted instances by LCR-Rot-hop from the SemEval-2016 training set. The subsets are sampled with replacement. We later report the averages and standard deviations of the accuracies obtained with 10 diagnostic classifiers, each trained on one of the 10 considered subsets.

\begin{figure*}[t!]
	\begin{minipage}[c]{0.48\textwidth}
	   \centering
	   \includegraphics[width=7.5cm]{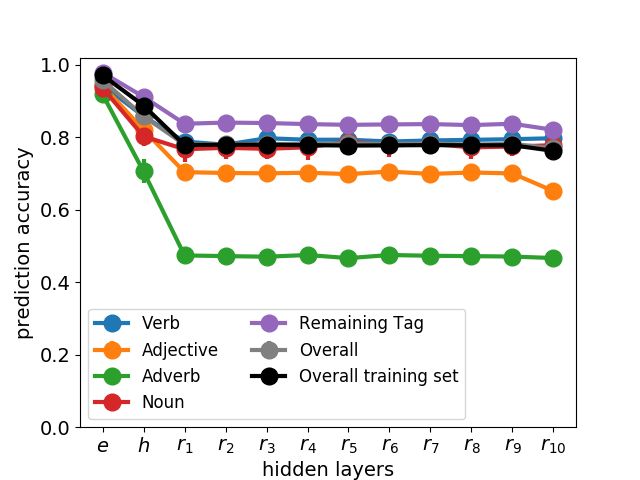}
	   \subcaption{Correctly predicted test set}
	\end{minipage}
	\begin{minipage}[c]{0.48\textwidth}
	   \centering
	   \includegraphics[width=7.5cm]{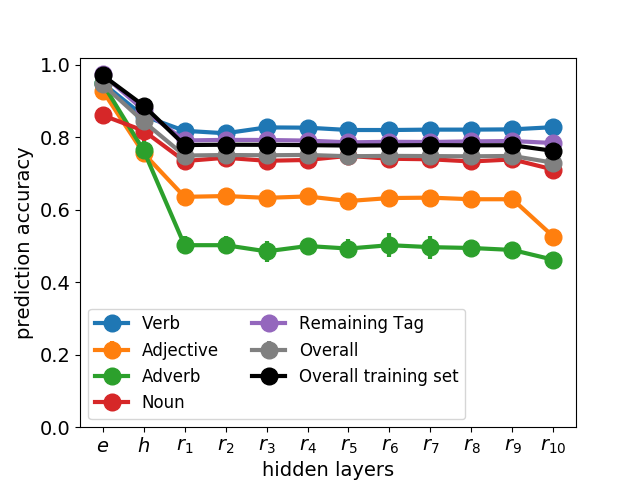}
	   \subcaption{Incorrectly predicted test set}
	\end{minipage}
	\caption{Accuracies of predicting the \textit{part-of-speech}}
	\label{fig:acc_pos}
	\vspace{-3mm}
\end{figure*}

\begin{figure*}[t!]
	\begin{minipage}[c]{0.48\textwidth}
	   \centering
	   \includegraphics[width=7.5cm]{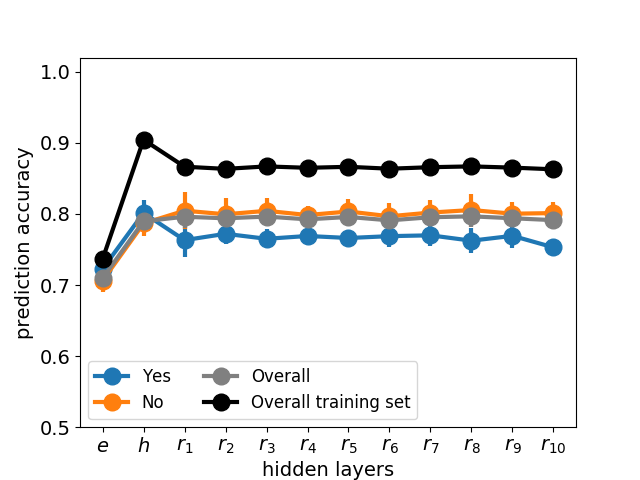}
	   \subcaption{Correctly predicted test set}
	\end{minipage}
	\begin{minipage}[c]{0.48\textwidth}
	   \centering
	   \includegraphics[width=7.5cm]{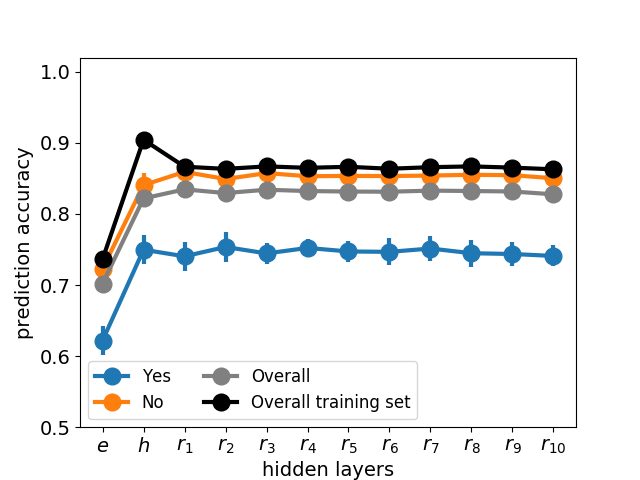}
	   \subcaption{Incorrectly predicted test set}
	\end{minipage}
	\caption{Accuracies of predicting the \textit{presence of a relation with an aspect}}
	\label{fig:acc_rel}
\end{figure*}

Inspired by the previous work \cite{DBLP:conf/acl/BelinkovDDSG17}, all our diagnostic classifiers are simple neural networks with one hidden layer of 300 neurons with a ReLU activation function. The output is determined by a dense layer with a softmax function. We initialise the weights by a uniform distribution $U(-0.1, 0.1)$ and the biases by zero for each layer. We use a learning rate of 0.0001 and a batch size of 20. The number of epochs is 90 except for the diagnostic classifiers predicting the \textit{part-of-speech}. Since these classifiers tend to end up in a saddle point after too many epochs, we set their number of epochs to 20. 

After we train all diagnostic classifiers on the internal layers and class labels of the randomly selected words in the correctly predicted opinions from the SemEval-2016 training dataset, we predict the class labels from the internal layers ($e_i, h_i, r_{i1}, ..., r_{in}$) of all words in the SemEval-2016 test dataset, and verify the accuracy of the diagnostic classifiers. If a diagnostic classifier predicts the class labels with a high accuracy, we conclude that the corresponding hidden layer is keeping track of the corresponding information.

\begin{figure*}[h!]
	\begin{minipage}[c]{0.48\textwidth}
	   \centering
	   \includegraphics[width=7.5cm]{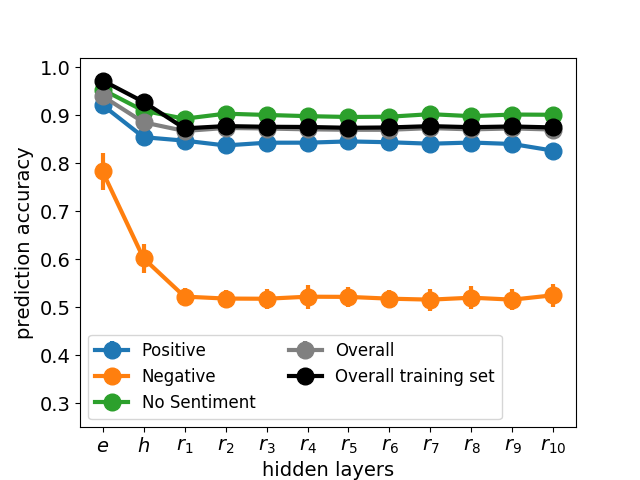}
        \subcaption{Correctly predicted test set}
	\end{minipage}
	\begin{minipage}[c]{0.48\textwidth}
	   \centering
	   \includegraphics[width=7.5cm]{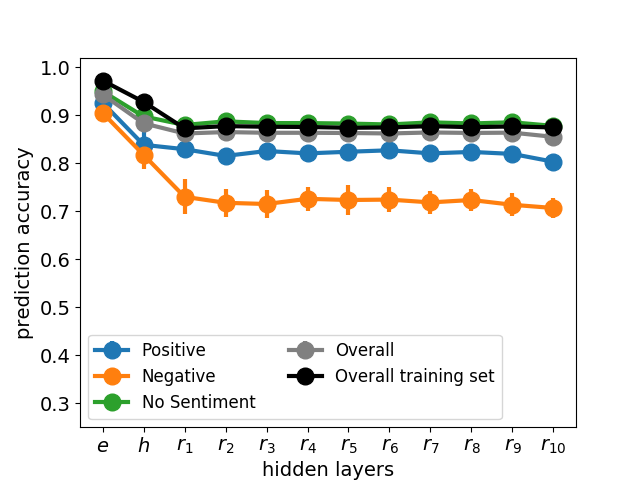}
	   \subcaption{Incorrectly predicted test set}
	\end{minipage}
	\caption{Accuracies of predicting the \textit{sentiment value}}
	\label{fig:acc_sent}
	\vspace{-3mm}
\end{figure*}

\begin{figure*}[h!]
	\begin{minipage}[c]{0.48\textwidth}
	   \centering
	   \includegraphics[width=7.5cm]{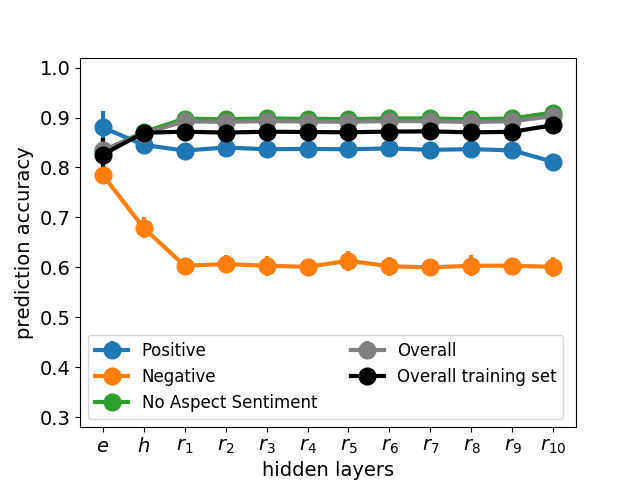}
	   \subcaption{Correctly predicted test set}
	\end{minipage}
	\begin{minipage}[c]{0.48\textwidth}
	   \centering
	   \includegraphics[width=7.5cm]{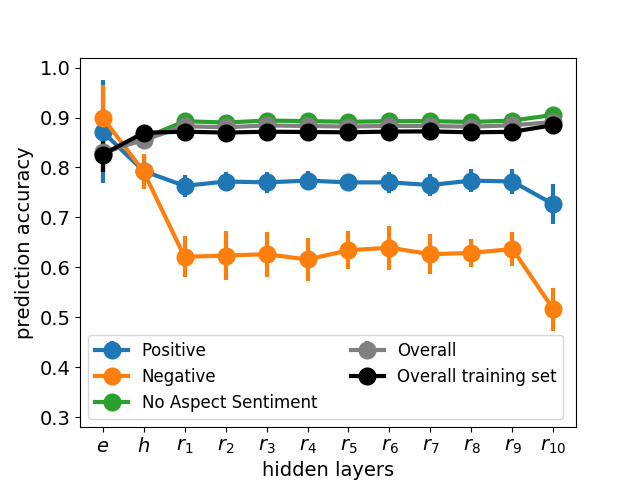}
	   \subcaption{Incorrectly predicted test set}
	\end{minipage}
	\caption{Accuracy of predicting the \textit{aspect-related sentiment value}}
	\label{fig:acc_as_sent}
\end{figure*}

\section{Evaluation}\label{sec:evaluation}

The following Section \ref{sec:performance_model} and Section \ref{sec:performance_diagnostic} present the performances of the LCR-Rot-Hop neural network and diagnostic classifiers, respectively.

\subsection{Performance of the LCR-Rot-hop} \label{sec:performance_model}
Instead of using 3 hops like \cite{Wallaart2018}, we train the LCR-Rot-hop model using 10 hops that allows us to better understand the information detection process of the rotatory attention. The model runs using 150 epochs, and a learning rate of 0.05, an $L_2$-norm regularisation term of 0.00001, a momentum of 0.9, and a batch size of 20. Furthermore, during training 50\% of neurons are randomly dropped out of the network. Table \ref{tab:lcr_rot_hop} gives its accuracy results on the SemEval-2016 training and test data for a single run. Unlike the \citeauthor{Wallaart2018}'s work, we do not run the model in two steps, which means that the LCR-Rot-hop model runs over the entire test set, without applying the ontology first. This is because we are interested to measure the performance of the neural network only.

\begin{table}[t!]
\caption{Accuracy results of the LCR-Rot-hop (with 10 hops)}
\centering
\footnotesize
\begin{tabular}{l c c c c}
\toprule
& \multicolumn{2}{c}{{Training set}} & \multicolumn{2}{c}{{Test set}}\\
\cmidrule(lr){2-3} \cmidrule(lr){4-5}
& {Freq.} & {acc. (\%)} & {Freq.} & {acc. (\%)}\\
\midrule Positive & 1319 & 99.4 & 483 & 94.6 \\
Neutral & 72 & 95.8 & 32 & 25.0\\
Negative & 488 & 95.9 & 135 & 69.6\\
Overall & 1879 & 98.4 & 650 & 86.0\\
\bottomrule
\end{tabular}
\label{tab:lcr_rot_hop}
\vspace{-4mm}
\end{table}

\subsection{Performance of the diagnostic classifiers} \label{sec:performance_diagnostic}
Figures \ref{fig:acc_pos}, \ref{fig:acc_rel}, \ref{fig:acc_sent}, and \ref{fig:acc_as_sent} illustrate the mean accuracies, and standard deviations of predicting the \textit{part-of-speech}, \textit{sentiment value}, \textit{presence of a relation with an aspect}, and \textit{aspect-related sentiment value} of the words in the correctly and incorrectly predicted opinions in the test data set. Since the standard deviations are rather small, the vertical lines, which represent the standard deviations in the graphs, are fairly short, and often not observable. We plot a colored line, for the prediction performance for each class label over the test set. We plot a grey line for the overall mean predicting performance over the test set, and we plot a black line for the overall mean prediction performance over the correctly predicted opinions from the training set. The information is split over over the internal representations of LCR-Rot-hop model: embeddings ($e$), hidden states ($h$), and context representation ($r$).

The performance of predicting the \textit{part-of-speech} by its diagnostic classifiers is illustrated by Figure \ref{fig:acc_pos}. Given that the information about the performance is split between the internal layers, it allows us to notice the much higher accuracy of predicting the \textit{part-of-speech} from the embeddings than of predicting from the context representations. Furthermore, the internal layers of the correctly predicted test set determine the class label ``Adjective'' much better than the same layers of the incorrectly predicted test set. The reason behind this lies in the fact that the adjectives have usually \textit{sentiment value}, and affect the overall sentiment orientation.

The performance of predicting the \textit{presence of a relation} with the aspect by its diagnostic classifiers is illustrated by Figure \ref{fig:acc_rel}. Unlike the case of \textit{part-of-speech}, we can predict the absence and the presence of a relation better from the context representations than from the embeddings. This presumption is valid, especially in predicting the absence of a relation. Remarkably, the internal layers of the incorrectly predicted test set indicate much better the absence of a relation than the internal layers of the correctly predicted test set. 

Figure \ref{fig:acc_sent} illustrates the performance of predicting the \textit{sentiment value} by its diagnostic classifiers. Analyzing the figure, we notice that the classifiers applied at the level of internal layers are particularly good at predicting ``No Sentiment'', and poor at predicting ``Negative''. Furthermore, from Figure \ref{fig:acc_sent} we see that it is easier to extract the sentiment value from the embeddings than from the context representations. Also, we notice that the correctly predicted test set indicates better the words with the label ``No Sentiment'' than the scenario with incorrect predictions. Knowing which words in the opinion do not have a defined sentiment value, may be more important for correctly predicting the sentiment value, than knowing the sentiment words.

Given Figure \ref{fig:acc_as_sent}, we see that the context representation predicts slightly better the \textit{aspect-related sentiment value} than the embeddings. Also, the diagnostic classifiers identify more easily the words with no sentiment value or unrelated to aspects. In addition, Figure \ref{fig:acc_as_sent} shows the label ``Positive'' from the internal layers of the correctly predicted test set is predicted with a higher accuracy than from the case with incorrect predictions. Probably, keeping track of positive words, which are related to the aspect, has a greater effect on predicting the sentiment value correctly than keeping track of negative words, since Table \ref{tab:lcr_rot_hop} shows us that the model predicts positive opinions much more accurately than negative opinions.

Looking at all four figures, we see that we do not predict more accurately from the context representation of the first round than from the context representation of the tenth round, or vice versa. Their prediction accuracies are in general about the same value. Apparently, iterating the rotatory attention mechanism does not contribute to an improvement or deterioration in the prediction accuracies of the diagnostic classifiers. The context representations are keeping track of roughly the same information. While the context representation are keeping track of roughly the same information, the embeddings representations and hidden states bring informational diversity. 

\section{Conclusion} \label{sec:concludingremarks}
\label{sec:conclusion}
In this paper, we propose various diagnostic classifiers in order to examine whether the ABSC state-of-the-art neural attention model LCR-Rot-hop encodes the \textit{part of speech}, the \textit{presence of aspect relation}, the \textit{sentiment value}, and the \textit{aspect-related sentiment value}, deemed as useful information at the word level for the considered task. We conclude that the context representation determines mostly the \textit{presence of a relation with an aspect}, and the \textit{aspect-related sentiment value}, without having a relevant performance difference among the 10 considered context representations. On the other hand, the remaining two features \textit{part-of-speech}, and \textit{sentiment value} are better detected by the embedding representations.


As for future work, we advice to optimise the hyper-parameters of the LCR-Rot-hop model, and the diagnostic classifiers. Moreover, we suggest to better specify the \textit{sentiment value}, and the \textit{presence of a relation towards the aspect}, since the ontology does not contain all relevant words, and the sentiment score from NLTK SentiWordNet, which is based on the most frequently used form, might not be the correct sentiment score. Furthermore, we propose to compute the prediction accuracy only over the words that indicate the sentiment value of the opinion, instead of computing the prediction accuracy over all words. In doing so, we can better understand how these words affect the final sentiment classification of the LCR-Rot-hop model. As well, we consider important to check the applicability of our inferences also on other sentiment datasets that comprise not only restaurant opinions.

\bibliographystyle{ACM-Reference-Format}
\balance
\bibliography{bibliography}

\end{document}